\documentclass[sn-mathphys-num]{sn-jnl}

\usepackage{amsmath}
\usepackage{amsfonts}
\usepackage{amssymb}
\usepackage{graphicx}
\usepackage{enumitem}
\usepackage{cleveref}
\usepackage{bm}
\usepackage{url}
\usepackage{tikz}
\usetikzlibrary{positioning}
\usetikzlibrary{shapes.geometric}
\usepackage{circuitikz}
\usepackage{algorithm}
\usepackage{algpseudocode,algorithm}
\algnewcommand\algorithmicinput{\textbf{Input:}}
\algnewcommand\Input{\item[\algorithmicinput]}
\algnewcommand\algorithmicoutput{\textbf{Output:}}
\algnewcommand\Output{\item[\algorithmicoutput]}

\algnewcommand\algorithmicforeach{\textbf{for each}}
\algdef{S}[FOR]{ForEach}[1]{\algorithmicforeach\ #1\ \algorithmicdo}

\makeatletter
\AtBeginDocument{%
  \def\doi#1{\url{https://doi.org/#1}}}
\makeatother

\newcommand{\R}[0]{\mathbb{R}}

\newcommand{\ideal}[1]{{\langle{#1}\rangle}}

\begin{document}

\title[An Optimized Path Planning of Manipulator Using Spline Curves]{An Optimized Path Planning of Manipulator Using Spline Curves and Real Quantifier Elimination Based on Comprehensive Gr\"obner Systems}

\author[1]{\fnm{Yusuke} \sur{Shirato}} %\email{}
\author[1]{\fnm{Natsumi} \sur{Oka}} %\email{}
\author*[2]{\fnm{Akira} \sur{Terui}}\email{terui@math.tsukuba.ac.jp}
\author[3]{\fnm{Masahiko} \sur{Mikawa}}\email{mikawa@slis.tsukuba.ac.jp}

\affil[1]{
  \orgdiv{Graduate School of Pure and Applied Sciences},
  \orgname{University of Tsukuba},
  \orgaddress{\city{Tsukuba-shi}, \postcode{305-8571}, \state{Ibaraki}, \country{Japan}}%
}

\affil*[2]{
  \orgdiv{Faculty of Pure and Applied Sciences},
  \orgname{University of Tsukuba},
  \orgaddress{\city{Tsukuba-shi}, \postcode{305-8571}, \state{Ibaraki}, \country{Japan}}%
}

\affil[3]{
  \orgdiv{Faculty of Library, Information and Media Science},
  \orgname{University of Tsukuba},
  \orgaddress{\city{Tsukuba-shi}, \postcode{305-8550}, \state{Ibaraki}, \country{Japan}}%
}

% \author{Yusuke Shirato}
% \address{
%   Graduate School of Pure and Applied Sciences \\
%   University of Tsukuba \\
%   Tsukuba-shi, Ibaraki 305-8571 \\
%   Japan
% }

% \author{Natsumi Oka}
% \address{
%   Graduate School of Pure and Applied Sciences \\
%   University of Tsukuba \\
%   Tsukuba-shi, Ibaraki 305-8571 \\
%   Japan
% }

% \author{Akira Terui}
% \address{
%   Faculty of Pure and Applied Sciences \\
%   University of Tsukuba \\
%   Tsukuba-shi, Ibaraki 305-8571 \\
%   Japan
% }
% \email{terui@math.tsukuba.ac.jp}

% \author{Masahiko Mikawa}
% \address{
%     Faculty of Library, Information and Media Science \\
%     University of Tsukuba \\
%     Tsukuba-shi, Ibaraki 305-8550 \\
%     Japan
% }
% \email{mikawa@slis.tsukuba.ac.jp}


% \begin{abstract}
  \abstract{This paper presents an advanced method for addressing the inverse kinematics and optimal path planning challenges in robot manipulators. The inverse kinematics problem involves determining the joint angles for a given position and orientation of the end-effector. Furthermore, the path planning problem seeks a trajectory between two points. Traditional approaches 
  in computer algebra have utilized Gr\"obner basis computations to solve these problems, offering a global solution but at a high computational cost. To overcome the issue, the present authors have proposed a novel approach that employs the Comprehensive Gr\"obner System (CGS) and CGS-based quantifier elimination (CGS-QE) methods to efficiently solve the inverse kinematics problem and certify the existence of solutions for trajectory planning. This paper extends these methods by incorporating smooth curves via cubic spline interpolation for path planning and optimizing joint configurations using shortest path algorithms to minimize the sum of joint configurations along a trajectory. This approach significantly enhances the manipulator's ability to navigate complex paths and optimize movement sequences.}
  % \keywords{Comprehensive Gr\"obner Systems \and Robotics \and Inverse kinematics \and 
  % Cubic spline interpolation \and Shortest path algorithms}
% \end{abstract}

\maketitle 

\section{Introduction}
\label{sec:introduction}

This paper discusses a method for solving the inverse kinematics problem and the optimal path planning problem for a robot manipulator.
A manipulator is a robot with links corresponding to human arms and joints corresponding to human joints, and the tip is called the end-effector.
The inverse kinematics problem for manipulators is to find the angle of each joint, given the position and orientation of the end-effector.
The path planning problem is to find a path to move the end-effector between two specified positions \cite{sic-kha2016}.

When operating the manipulator, one needs to solve the inverse kinematics problem (or the path planning problem, respectively) for the desired end-effector position (or the series of positions, respectively).

Methods of solving inverse kinematics problems for manipulators by reducing the inverse kinematics problem to a system of polynomial equations and using the Gr\"obner basis has been proposed \cite{fau-mer-rou2006,kal-kal1993,ric-sch-ces2021,uch-mcp2011,uch-mcp2012}.
Solving the inverse kinematics problem using the Gr\"obner basis computation has an advantage that the global solution of the inverse kinematics problem can be obtained before the end-effector will actually be ``moved'' by simulation or other means.
On the other hand, the Gr\"obner basis computation has the disadvantage of relatively high computational cost compared to local solution methods such as the Newton method.
Furthermore, when solving a path planning problem using the Gr\"obner basis computation, it is necessary to solve the inverse kinematics problem for each point on the path, which is even more computationally expensive.

The third and fourth authors have also previously proposed methods for solving inverse kinematics problems using Gr\"obner basis computations 
\cite{hor-ter-mik2020,ota-ter-mik2021,yos-ter-mik2023}.
The authors' contributions in their previous work \cite{yos-ter-mik2023} are as follows.
We have proposed a method for solving the inverse kinematics problem and the trajectory planning problem of a 3 Degree-Of-Freedom (DOF) manipulator using the Comprehensive Gr\"obner System (CGS) \cite{wei1992}.
In the proposed method, the inverse kinematic problem has been expressed as 
a system of polynomial equations with the coordinates of the end-effector as parameters
and the CGS is calculated in advance to reduce the cost of Gr\"obner basis computation
for each point on the path.
In addition, we have proposed a method for solving the inverse kinematics problems using the CGS-QE method \cite{fuk-iwa-sat2015}, which is a quantifier elimination (QE) method based on CGS computations, to certify the existence of a (real) solution.
Furthermore, we have proposed a method to certify the existence of a solution to the whole trajectory planning problem using the CGS-QE method, where the points on the trajectory are represented by parameters.

This paper proposes the following method as an extension of the previous work \cite{yos-ter-mik2023}.
\begin{enumerate}
  \item Extension of paths used in path planning problems: while the previous work has used straight lines, this paper uses smooth curves generated by the cubic spline interpolation passing through given points.
  For example, a curved path allows us to plan paths that avoid obstacles.
  \item Optimization of the joint configuration obtained as the solution to the path planning problem: when solving the path planning problem, there can be multiple solutions to the inverse kinematics problem at each point on the path. 
  In this case, the question is which of the solutions for adjacent points can be connected to minimize the sum of the configurations of the entire sequence of the joints.
  In this paper, we reduce this problem to the shortest path problem of a weighted graph and propose a method to compute the optimal sequence of joint configurations using shortest path algorithms.
\end{enumerate}

This paper is organized as follows.
In \Cref{sec:inverse-kinematics}, we describe the inverse kinematics problem for a 3-DOF manipulator and the method of solving it.
In \Cref{sec:spline-interpolation}, the method of generating trajectories using cubic spline interpolation is proposed.
In \Cref{sec:optimal-path}, we describe the method of optimizing the joint configuration obtained as the solution to the path planning problem.
In \Cref{sec:conclusion}, we give concluding remarks and future research directions.

% \section{Solving path planning problems in 3-DOF manipulators}
% \label{sec:inverse-kinematics}

\section{Solving the inverse kinematic problem}
\label{sec:inverse-kinematics}

The manipulator used in this paper is myCobot 280 \cite{mycobot-280} from Elephant Robotics, Inc.
(hereafter referred to simply as ``myCobot'').
Although myCobot is a 6-DOF manipulator, we treat it as a 3-DOF manipulator by operating only the three main joints used to move the end-effector while the remaining joints are fixed.

% \subsection{Formulation of the Forward kinematics problem}

\begin{figure}
  \centering
  \includegraphics[scale=0.3]{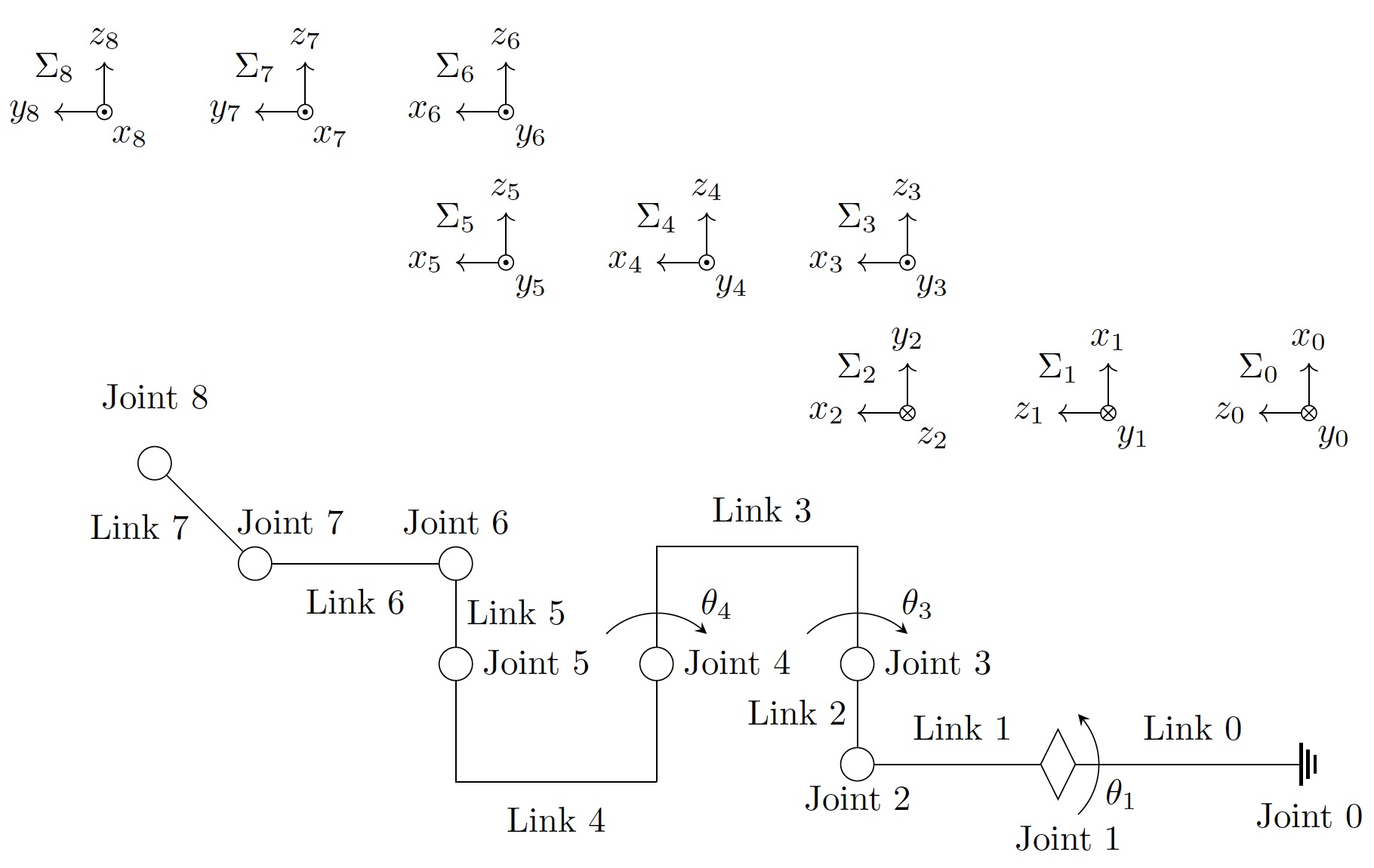}
  \caption{The model of myCobot.}
  \label{fig:mycobot}
\end{figure}

The forward kinematics problem for myCobot is derived using the modified Denavit-Hatenberg 
convention (hereafter abbreviated as ``D-H convention''), which is standard in robotics \cite{sic-kha2016}.
Each link is sequentially named Link 0, Link 1, \dots, and Link 7, starting from a fixed point to the end effector. 
For $i=1,2,\dots,7$, the joint connecting Link $(i-1)$ and Link $i$ is called Joint $i$. 
Additionally, the endpoint of Link 0 that is not connected to Joint 1 is called Joint 0, and the 
end effector is called Joint 8. 
The 3D right-handed coordinate system at joint $i$ ($i=0,1,\dots,8)$ is denoted by $\Sigma_i$. 
The origin of $\Sigma_1$ is Joint 2, and for $i \ne 1$, the origin of $\Sigma_i$ is Joint $i$. If Joint $i$ is a rotational joint, let the $z_i$ axis be its rotation axis. Other axes and their directions are set to make the transformation matrix from coordinate system $\Sigma_i$ to coordinate system $\Sigma_{i-1}$ relatively simple.
\Cref{fig:mycobot} shows the model of myCobot with the coordinates of each joint. 
In the representation of coordinate axes, the axis pointing from the back of the page to the front is shown as the ``arrowhead,'' while the axis pointing from the front of the page to the back is depicted as the ``arrow tail'' (a circle with a cross mark).

Next, we provide the transformation matrix between the coordinate systems of myCobot. To represent the relationship between adjacent coordinate systems, we define the parameters $a_i$, $\alpha_i$, $d_i$, $\theta_i$ for $\Sigma_{i-1}$ and $\Sigma_i$ as follows:
\begin{itemize}
  \item $a_i$: The distance between the $z_{i-1}$ axis and the $z_i$ axis [mm],
  \item $\alpha_i$: The angle from the $z_{i-1}$ axis to the $z_i$ axis around the $x_{i-1}$ axis [rad],
  \item $d_i$: The distance between the $x_{i-1}$ axis and the $x_i$ axis [mm],
  \item $\theta_i$: The angle from the $x_{i-1}$ axis to the $x_i$ axis around the $z_i$ axis [rad].
\end{itemize}
\Cref{tab:pamameters} shows the parameters of each joint.
By using these parameters, the transformation matrix $^{i-1}T_i$ from $\Sigma_i$ to $\Sigma_{i-1}$ is given as
\begin{align}
    \label{eq:transformation-matrix}
    ^{i-1}T_i &=
    \left(
      \begin{array}{cccc}
        1 & 0 & 0 & a_i \\
        0 & 1 & 0 & 0 \\
        0 & 0 & 1 & 0 \\
        0 & 0 & 0 & 1
      \end{array}
    \right)
    \left(
      \begin{array}{cccc}
        1 & 0 & 0 & 0 \\
        0 & \cos \alpha_i & -\sin \alpha_i & 0  \\
        0 & \sin \alpha_i & \cos \alpha_i & 0 \\
        0 & 0 & 0 & 1
      \end{array}
    \right)
    \nonumber \\
    &\quad\times
    \left(
      \begin{array}{cccc}
        1 & 0 & 0 & 0 \\
        0 & 1 & 0 & 0 \\
        0 & 0 & 1 & d_i \\
        0 & 0 & 0 & 1
      \end{array}
    \right)
    \left(
      \begin{array}{cccc}
        \cos \theta_i & -\sin \theta_i & 0 & 0 \\
        \sin \theta_i & \cos \theta_i & 0 & 0 \\
        0 & 0 & 1 & 0 \\
        0 & 0 & 0 & 1 
      \end{array}
    \right) 
    \nonumber \\
    &= 
    \left(
      \begin{array}{cccc}
        \cos \theta_i & -\sin \theta_i & 0 & a_i \\
        \cos \alpha_i \sin \theta_i & \cos \alpha_i \cos \theta_i & -\sin \alpha_i & -d_i \sin \alpha_i   \\
        \sin \alpha_i \sin \theta_i & \sin \alpha_i \cos \theta_i & \cos \alpha_i & d_i \cos \alpha_i \\
        0 & 0 & 0 & 1
      \end{array}
    \right),
\end{align}
where the affine coordinates are used to represent the transformation matrix.

\begin{table}[t]
  \centering
  \caption{Parameters of each joint.}
  \label{tab:pamameters}
  \begin{tabular}{c|cccc}
    $i$	  & $a_i$	& $\alpha_i$ 		  	& $d_i$ 	& $\theta_i$  \\ 
    \hline
    1    	& 0     & 0 				        & 131.56 	& $\theta_1$  \\
    2    	& 0		  & $-\frac{\pi}{2}$ 	& 0 		  & $-\frac{\pi}{2}$ \\
    3    	& 0 		& $-\frac{\pi}{2}$ 	& 33.195	& $\theta_3 $ \\
    4    	& 110.4	& 0			 	          & 0	    	& $\theta_4$ \\
    5    	& 96.0 	& 0			 	          & 0 		  & 0 \\
    6    	& 0 		& 0			 	          & 33.195 	& 0 \\
    7    	& 73.18	& 0			 	          & 0	 	    & $-\frac{\pi}{2}$ \\  
    8    	& 43.6 	& 0 			        	& 0 		  & 0 \\
    \hline
  \end{tabular}  
\end{table}

Let the coordinates of the end effector in the $\Sigma_0$ coordinate system be $(x,y,z)$, and the transformation matrix from the $\Sigma_8$ coordinate system to the $\Sigma_0$ coordinate system be
$T = {}^{0}T_1 {}^{1}T_2 {}^{2}T_3 {}^{3}T_4 {}^{4}T_5 {}^{5}T_6 {}^{6}T_7 {}^{7}T_8.$
Since the end effector is located at the origin of the $\Sigma_8$ coordinate system, the coordinates of the end effector in the $\Sigma_0$ coordinate system can be expressed as
\begin{align}
  \label{eq:forward-kinematics}
  \left(
    \begin{array}{c}
      x \\ y \\ z \\ 1
    \end{array}
  \right)
  = T
  \left(
  \begin{array}{c}
    0 \\ 0 \\ 0 \\ 1
  \end{array}
  \right).
\end{align}
By expressing $s_i=\sin{\theta_i}$ and $c_i=\cos{\theta_i}$ for $i=1,3,4$, 
\cref{eq:forward-kinematics} can be rewritten as
\begin{equation}
  \label{eq:forward-kinematics-formula}
  \begin{split}
    -169.18s_1s_3c_4 -169.18s_1c_3s_4 -43.6s_1s_3s_4 +43.6s_1c_3c_4 \quad & \\
    -110.4s_1s_3 -66.39c_1 +x &= 0,\\
    169.18c_1s_3c_4 +169.18c_1c_3s_4 +43.6c_1s_3s_4 -43.6c_1c_3c_4 \quad  & \\
    +110.4c_1s_3 -66.39s_1 +y &= 0, \\
    -169.18c_3c_4 +169.18s_3s_4 -43.6c_3s_4 -43.6s_3c_4  \quad & \\
     -110.4c_3 - 131.56 +z&= 0, \\
    s_1^2 + c_1^2 - 1 = 0, \quad s_3^2 + c_3^2 - 1 = 0, \quad s_4^2 + c_4^2 - 1 &= 0.
  \end{split}
\end{equation}

In the following, for a set of polynomials $F=\{f_1,\dots,f_m\}\subset \R[c_1,s_1,c_3,s_3,c_4,s_4]$, The system of polynomial equations $f_1=\cdots=f_m=0$ is denoted by $F=0$.
Since the first, second, and third equations of \cref{eq:forward-kinematics-formula} contain decimal coefficients, both sides of these equations are actually multiplied by 100.
Thus, we obtain the inverse kinematic problem as the one to find the solution $c_1,s_1,c_3,s_3,c_4,s_4$ to the system of polynomial equations $F=0$ with $F=\{f_1,\dots,f_6\}$, where
\begin{equation}
  \label{eq:inverse-kinematics-formula}
  \begin{split}
  f_1&=-16918s_1s_3c_4 -16918s_1c_3s_4 -4360s_1s_3s_4 +4360s_1c_3c_4 -11040s_1s_3\\
  &\quad -6639c_1 +100x,\\
  f_2&=16918c_1s_3c_4 +16918c_1c_3s_4 +4360c_1s_3s_4 -4360c_1c_3c_4 +11040c_1s_3\\
  &\quad -6639s_1 +100y,\\
  f_3&=-16918c_3c_4 +16918s_3s_4 -4360c_3s_4 -4360s_3c_4 -11040c_3\\
  &\quad - 13156 +100z,\\
  f_4&=s_1^2 + c_1^2 - 1,\quad f_5=s_3^2 + c_3^2 - 1,\quad f_6=s_4^2 + c_4^2 - 1.
  \end{split}
\end{equation}
Note that the system of equations $F=0$ has parameters $x,y,z$ in the coefficients.

The inverse kinematic problem is solved as follows.
With our previously proposed method \cite[Algorithm 1]{yos-ter-mik2023}, 
before solving $F=0$, we calculate the CGS of $\ideal{f_1,\dots,f_6}$.
Then, for a given end-effector coordinate $(x,y,z)=(\alpha,\beta,\gamma)\in\R^3$,
determine the feasibility of such a configuration using the CGS-QE method.
If the configuration is feasible, we solve $F=0$ numerically after substituting parameters $x,y,z$ with $\alpha,\beta,\gamma$, respectively, to obtain the inverse kinematic solution.

\section{Path and trajectory planning using spline interpolation}
\label{sec:spline-interpolation}

% \subsection{Generating trajectories using cubic spline interpolation}

In the path planning problem, a finite number of points through which the path passes are given in advance, and a trajectory is generated on the smooth path passing through these points.
While a straight line was used as the path in our previous study, 
a path is generated using cubic spline interpolation \cite{far2002} in the present paper.

\subsection{Path planning using cubic spline interpolation}

Let $(x_0,y_0,z_0)$, $(x_1,y_1,z_1)$, $(x_2,y_2,z_2)$, $(x_3,y_3,z_3)$ be four different points given in $\R^3$.
For $j=0,1,2$, calculate a curve $C_j$ passing through 
$(x_j,y_j,z_j)$ and $(x_{j+1},y_{j+1},z_{j+1})$
in the form of a function $(X_j(s),Y_j(s),Z_j(s))$ with $s\in[0,1]$ as a parameter.
In cubic spline interpolation, $X_j(s),Y_j(s),Z_j(s)$ are cubic polynomials
given by
\begin{equation}
  \label{eq:qubic-spline}
  \begin{split}
    X_j(s) &= a_{X,j}s^3 + b_{X,j}s^2 + c_{X,j}s + d_{X,j}, \\
    Y_j(s) &= a_{Y,j}s^3 + b_{Y,j}s^2 + c_{Y,j}s + d_{Y,j}, \\
    Z_j(s) &= a_{Z,j}s^3 + b_{Z,j}s^2 + c_{Z,j}s + d_{Z,j},
  \end{split}
\end{equation}
where $a_{X,j},b_{X,j},...,d_{Z,j}$ are real numbers.
Furthermore, we impose the following conditions on these functions.
\begin{equation}
  \label{eq:spline-interpolation}
  % \small
  \begin{split}        
    & (X_j(j),Y_j(j),Z_j(j))=(x_j,y_j,z_j),\quad j=0,1,2,\\
    & (X_j(j+1),Y_j(j+1),Z_j(j+1)) \\
    &\quad =(x_{j+1},y_{j+1},z_{j+1}),\quad j=0,1,2,\\
    &(X'_j(j+1),Y'_j(j+1),Z'_j(j+1)) \\
    &\quad =(X'_{j+1}(j+1),Y'_{j+1}(j+1),Z'_{j+1}(j+1)),\quad j=0,1,\\
    &(X''_j(j+1),Y''_j(j+1),Z''_j(j+1)) \\
    &\quad =(X''_{j+1}(j+1),Y''_{j+1}(j+1),Z''_{j+1}(j+1)),\quad j=0,1.
  \end{split}
\end{equation}
In this paper, in addition to the conditions in \cref{eq:spline-interpolation},
we find the \emph{natural} cubic Spline interpolation that satisfies
\begin{equation}
  \label{eq:natural-spline}
  X''_0(0)=X''_3(3)=Y''_0(0)=Y''_3(3)=Z''_0(0)=Z''_3(3)=0.
\end{equation}

The coefficients of $X_j(s)$ are determined as follows (the coefficients of $Y_j(s)$ and $Z_j(s)$ are determined similarly).
In general, suppose that $N$ points $(x_0,y_0),\dots,(x_N,y_N)$ are given, and, 
for $j=0,\dots,N-1$, the curve $C_j$ passing through $(x_j,y_j,z_j)$ and $(x_{j+1},y_{j+1},z_{j+1})$ are to be calculated. 
Let $v_{x,j+1}=6(x_j-2x_{j+1}+x_{j+2})$ for $j=0,\dots,N-2$ and $v_{x,0}=v_{x,N-1}=0$.
Then, define $u_{X,j}$ as the solution to the following system of linear equations:
\begin{align}
  \label{eq:spline-interpolation-linear-equation}
  \begin{pmatrix}
    4 & 1 \\
    1 & 4 & 1 \\
      & \ddots & \ddots & \ddots \\
      &        & 1      & 4      & 1 \\
      &        &        & 1      & 4 \\
  \end{pmatrix}
  \begin{pmatrix}
    u_{X,1} \\ u_{X,2} \\ \vdots \\ u_{X,N-2} \\ u_{X,N-1}
  \end{pmatrix}
  =
  \begin{pmatrix}
    v_{x,1} \\ v_{x,2} \\ \vdots \\ v_{x,N-2} \\ v_{x,N-1} 
  \end{pmatrix}
  .
\end{align}
Furthermore, set $u_{X,0}=u_{X,N}=0$ by the natural spline condition (\cref{eq:natural-spline}).
Then, the coefficients $a_{X,j},b_{X,j},c_{X,j},d_{X,j}$ of $X_j(s)$ are determined as
\begin{align*}
  a_{X,j}&=\frac{1}{6}(u_{X,j+1}-u_{X,j}), & 
  b_{X,j}&=\frac{u_{X,j}}{2},\\
  c_{X,j}&=x_{j+1}-x_j-\frac{1}{6}(u_{X,j+1}+2u_{X,j}), &
  d_{X,j}&=x_j,
\end{align*}
for $j=0,\dots,N-1$.

\subsection{Solving the inverse kinematics problem}

We generate the trajectory for the curve $C_j$ obtained in the previous section by solving the inverse kinematics problem using the following procedure.

First, using the CGS-QE method, we determine the existence of solutions to the inverse kinematics problem for each curve $C_j$ ($j=0,1,2$) that constitutes the path to move the end-effector.
In \cref{eq:inverse-kinematics-formula}, replace $x,y,z$ in each equation by $X_j(s),Y_j(s),Z_j(s)$, respectively, to obtain a system of polynomial equations with $s$ as parameter. Let 
the result be $\bar{F}_j(s)$.

Next, we apply the CGS-QE method to $\bar{F}_j(s)$ to determine whether the system of equations $\bar{F}_j(s)=0$ has real solutions within the parameter $s\in[0,1]$.
If $\bar{F}_j(s)=0$ has real solutions within the parameter $s\in[0,1]$,
%  (i.e., within $s\in[0,3]$ throughout for $s$), 
we calculate the trajectory of the end-effector's position 
by changing $s$ with a series of time steps $t=0,\dots,T$, where $T$ is a positive integer.
For time $t=0,\dots,T$, let $s=f(t)$ where $f$ is a continuous function of $[0,T]\to[0,1]$.
To ensure that the end-effector has no velocity and acceleration at the beginning and end of the trajectory, we obtain $f$ as a polynomial of the smallest degree possible to satisfy $f'(t)=f''(t)=0$ at $t=0$ and $T$.
Then, we obtain 
\begin{equation}
  \label{eq:ft}
  f(t)=\left(\frac{6t^5}{T^5}-\frac{15t^4}{T^4}+\frac{10t^3}{T^3}\right)   
\end{equation}
(see \cite{lyn-par2017,yos-ter-mik2023}).

Finally, 
% we solve the inverse kinematics problem $t=0,\dots,T$. 
For $j=0,1,2$ and $t=0,\dots,T$, the configuration of the joints is obtained by solving the system of polynomial equations 
\begin{equation}
  \label{eq:inverse-kinematics-each-time}
  \bar{F}_j(f(t))=0.    
\end{equation}

\subsection{An algorithm for path and trajectory planning with spline interpolation}

Summarizing the above, an algorithm for path and trajectory planning with spline interpolation is 
given in \Cref{alg:spline-trajectory}.

\begin{algorithm}[t]
  \caption{Path and trajectory planning with spline interpolation}
  \label{alg:spline-trajectory}
  \begin{algorithmic}[1]
    \Input{The points that the end-effector of myCobot should pass through, 
    including the initial position and the target position: $(x_0,y_0,z_0),\dots,(x_N,y_N,z_N)$,
    The number of time steps: $T$}
    \Output{A set of the tuple $(\theta_1,\theta_3,\theta_4)$ for each of the $(N\times T+1)$ points on the spline curve passing through the input points}
    \State{$A\gets\emptyset$;}  
    \State{Solve the inverse kinematic problem \cref{eq:inverse-kinematics-formula} for 
    $(x,y,z)=(x_0,y_0,z_0)$ with the algorithm for the inverse kinematics problem 
    \cite[Algorithm 1]{yos-ter-mik2023};}
    \If{no solution is obtained}
      \State{\Return Error;}
    \EndIf
    \State{$A\gets A\cup\{(\theta_1,\theta_3,\theta_4)\}$;}
    \State{For $(x_0,y_0,z_0),\dots,(x_N,y_N,z_N)$, input to the algorithm for spline interpolation \cite{far2002} to obtain the spline curve $(X_j(s),Y_j(s),Z_j(s))$;}
    \For{$j\in[0,\dots,N-1]$}
      \State{Put $f(t)$ in \cref{eq:ft} into $s$ in $(X_j(s),Y_j(s),Z_j(s))$;}
      \For{$t\in[1,\dots,T]$}
        \State{Solve the inverse kinematic problem \cref{eq:inverse-kinematics-each-time} for $(X_j(t),Y_j(t),Z_j(t))$ with the algorithm for the inverse kinematics problem 
        \cite[Algorithm 1]{yos-ter-mik2023};}
        \If{no solution is obtained}
          \State{\Return Error, ($j, t$);}
        \Else
          \State{$A\gets A\cup\{(\theta_1,\theta_3,\theta_4)\}$;}
        \EndIf
      \EndFor
    \EndFor 
  \end{algorithmic}
\end{algorithm}

\subsection{Implementation and computational results}
\label{sec:spline-implementation}

This section describes the implementation and experimental results of the proposed method.
We implemented \Cref{alg:spline-trajectory} on the computer algebra system Risa/Asir \cite{nor-tak1992} and used it together with our existing implementation of inverse kinematics calculations \cite{yos-ter-mik2023}.
For the CGS calculation, we used an algorithm by Kapur et al. \cite{kap-sun-wan2010} with an implementation by Nabeshima \cite{nab2018}.
The existence of the root of the inverse kinematics problem by the CGS-QE method was verified using Risa/Asir with accompanying Wolfram Mathematica 13.1 \cite{mathematica13} for simplifying the expression.

We measured the computation time for the path and the trajectory planning of myCobot for each of the six types of input points $(x_0,y_0,z_0)$, $(x_1,y_1,z_1)$, $(x_2,y_2,z_2)$, $(x_3,y_3,z_3)$, as below, for $T=10$ and $T=50$.
\begin{description}
  \item[Test 1] $(-100,-100,100),(0,-150,50),(50,-100,50),(100,0,0)$,
  \item[Test 2] $(-150,-100,150),(0,-100,100),(100,-50,50),(100,100,0)$,
  \item[Test 3] $(-250,0,0),(-150,0,50),(-150,100,150),(250,100,100)$,
  \item[Test 4] $(100,50,0),(50,100,150),(-150,150,100),(-150,50,50)$,
  \item[Test 5] $(100,100,-50),(50,100,0),(-100,100,50),(-150,-50,100)$,
  \item[Test 6] $(150,150,0),(50,50,100),(-50,-50,100),(-150,-150,50)$.
\end{description}
The computing environment is as follows:
Intel Xeon Silver 4210 3.2 GHz, 
RAM 256 GB,
Linux Kernel 5.4.0,
Risa/Asir Version 20230315.

\Cref{tab:computation-time-trajectory} shows the result of the experiment.
The CGS of the ideal generated by the polynomials in \cref{eq:inverse-kinematics-formula} used in the experiment was calculated in advance\footnote{The CGS was calculated in approximately 0.16 [s] on the following computing environment: Intel Celeron J4125 2.00 GHz, RAM 5.6 GB,
Ubuntu 22.04.3 LTS, Risa/Asir: version 20230315.}. 
The columns ``$T=10$'' and ``$T=50$'' show each test's computation time with $T=10$ and $50$, respectively.
The data labeled ``Fail'' means that during the computation, it was discovered that part of the generated spline does not lie within the feasible region of myCobot for the given sequence of points, and thus, the computation was aborted.
The computation results suggest that the computation time is approximately linearly proportional to $T$. Additionally, it indicates that the curve designed with the spline may exceed the feasible region of myCobot, and measures are needed to address such cases.

\begin{table}[t]
  \centering
  \caption{Computation time for path and trajectory planning of myCobot.}
  \label{tab:computation-time-trajectory}
  \begin{tabular}{c|cc} 
    \hline
    Test & $T=10$ [s] & $T=50$ [s] \\
    \hline
    1 & 1.293 & 7.836\\ 
    2 & 1.342 & 6.490\\ 
    3 & 1.296 & 8.049\\ 
    4 & 1.574 & 7.616\\ 
    5 & 1.265 & 7.155\\ 
    6 & Fail & Fail\\ 
    \hline
    Average & 1.354 & 7.429 \\ 
    \hline
  \end{tabular}
\end{table}

\section{Choosing an optimal path from multiple solutions of inverse kinematics}
\label{sec:optimal-path}

When executing the trajectory planning of myCobot, as shown in the previous section,
the inverse kinematics problem \cref{eq:inverse-kinematics-formula} may have multiple solutions at each time $t$.
When actually operating the manipulator, it is necessary to select one solution uniquely. Therefore, in this section, we propose and examine two methods for deriving the optimal path for the manipulator's operation.

\begin{figure}
  \centering
  \begin{tikzpicture}[scale=0.75]
    %点のプロット
    \foreach \Point in {
    (0,6.5), (2,6.5), (4,6.5), (6,6.5),
    (0,5), (2,5), (4,5), (6,5),
    (0,3.5), (2,3.5), (4,3.5), (6,3.5),
    (0,1.5), (2,1.5), (4,1.5), (6,1.5), 
    (0,0), (2,0), (4,0), (6,0)
    }{\node at \Point {\textbullet};}
    %名前
    \draw (-1,6.5) node[]{$p_1$};
    \draw (-1,5) node[]{$p_2$};
    \draw (-1,3.5) node[]{$p_3$};
    \draw (-1,1.5) node[]{$p_{n-1}$};
    \draw (-1,0) node[]{$p_n$};
    \draw (3,7.5) node[]{Four possible configurations of the joints};
    \draw (0,7) node[]{$1$};
    \draw (2,7) node[]{$2$};
    \draw (4,7) node[]{$3$};
    \draw (6,7) node[]{$4$};
    \draw (0,6.5) node[below right]{$p_{1,1}$};
    \draw (2,6.5) node[below right]{$p_{1,2}$};
    \draw (4,6.5) node[below right]{$p_{1,3}$};
    \draw (6,6.5) node[below right]{$p_{1,4}$($\theta_1^{(1,4)},\theta_3^{(1,4)},\theta_4^{(1,4)}$)};
    \draw (0,5) node[below right]{$p_{2,1}$};
    \draw (2,5) node[below right]{$p_{2,2}$};
    \draw (4,5) node[below right]{$p_{2,3}$};
    \draw (6,5) node[below right]{$p_{2,4}$};
    \draw (0,3.5) node[below right]{$p_{3,1}$};
    \draw (2,3.5) node[below right]{$p_{3,2}$};
    \draw (4,3.5) node[below right]{$p_{3,3}$};
    \draw (6,3.5) node[below right]{$p_{3,4}$};
    \draw (0,1.5) node[below right]{$p_{n-1,1}$};
    \draw (2,1.5) node[below right]{$p_{n-1,2}$};
    \draw (4,1.5) node[below right]{$p_{n-1,3}$};
    \draw (6,1.5) node[below right]{$p_{n-1,4}$};
    \draw (0,0) node[below right]{$p_{n,1}$};
    \draw (2,0) node[below right]{$p_{n,2}$};
    \draw (4,0) node[below right]{$p_{n,3}$};
    \draw (6,0) node[below right]{$p_{n,4}$};
    %…
    \fill[black](-1,2.2)circle(0.03);
    \fill[black](-1,2.5)circle(0.03);
    \fill[black](-1,2.8)circle(0.03);
    \fill[black](3,2.2)circle(0.03);
    \fill[black](3,2.5)circle(0.03);
    \fill[black](3,2.8)circle(0.03);
    %経路線
    \draw (6,6.5)--(6,5)--(2,3.5);
    \draw (2,1.5)--(0,0); 
  \end{tikzpicture}
  \caption{Four possible configurations of the joints at each point on the path.}
  \label{fig:four-possible-configurations}
\end{figure}
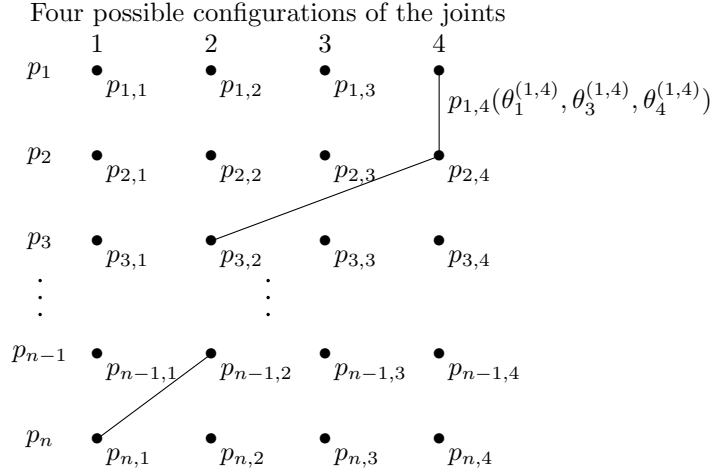

As a preparation, first, let the points where the end effector moves be $p_1$ (starting point), $p_2$, \dots, $p_{n-1}$, $p_n$ (ending point). 
The points $p_1,\dots,p_n$ divide the curve (derived with the proposed method) into $n-1$ equal segments when the end effector passes through.
In the previous section on path and trajectory planning, we observed four solutions to the inverse kinematics problem at each point on the trajectory.
This means that at each of $p_1,\dots,p_n$, there are four possible configurations of the joints, which we denote as $p_{1,1},p_{1,2},\dots,p_{1,4},\dots,p_{n,1},p_{n,2},\dots, p_{n,4}$.
In other words, $p_{i,j}$ ($i=1,\ldots,n$, $j=1,2,3,4$) refers to the $j$-th possible configuration of the joints at point $p_i$.
Additionally, since each $p_{i,j}$ has values for $\theta_1$, $\theta_3$, and $\theta_4$ (in units of [rad]) as its elements, we denote them as 
$p_{i,j}=(\theta_1^{(i,j)},\theta_3^{(i,j)},\theta_4^{(i,j)})$ (see \cref{fig:four-possible-configurations}).

Next, we define the magnitude of the joint variations between each point.
When the end effector moves from point and configuration (hereafter simply referred to as ``point'') $p_{i,j}$ to point $p_{i+1,k}$ ($k=1,2,3,4$), the magnitude of the joint variations is defined by 
\begin{equation}
  \label{eq:joint-variation}
  s_{i,j,i+1,k}=|\theta_1^{(i+1,k)}-\theta_1^{(i,j)}|+|\theta_3^{(i+1,k)}-\theta_3^{(i,j)}|
  +|\theta_4^{(i+1,k)}-\theta_4^{(i,j)}|.
\end{equation}

\subsection{Minimizing the magnitude of joint variations between each point}
\label{sec:optimal-path-between-points}

The first method is trajectory planning that minimizes the magnitude of joint variations 
$s_{i,j,i+1,k}$ between each point. The following calculations are repeated for $m=1,2,3,4$.
\begin{enumerate}[label=Step \arabic*., leftmargin=13mm]
  \item Calculate all $s_{1,m,2,k}$ from the starting point $p_{1,m}$ to the points $p_{2,k}$ ($k=1,2,3,4$), and select the point $p_{2,k}$ with the smallest $s_{1,m,2,k}$, denoting it as $p_{2,k'}$. \label{step:1}
  \item From the point ($p_{2,k'}$) selected in \cref{step:1}, calculate all $s_{2,k',3,k}$ to the points $p_{3,k}$, and select the point $p_{3,k}$ with the smallest $s_{2,k',3,k}$, denoting it as $p_{3,k'}$.
  \item Similarly, from the point ($p_{i,k'}$) selected in the previous step, calculate all $s_{i,k',i+1,k}$ to the next points ($p_{i+1,k}$), and select the point with the smallest $s_{i,k',i+1,k}$, denoting it as $p_{i+1,k'}$. Repeat this process until the endpoint $p_{n,k'}$ is determined.
\end{enumerate}
Among the paths starting from these four initial points, the path that minimizes the sum of the magnitudes of the joint variations between the selected points,
expressed as
\[
  S_{1,m,n,k'}=s_{1,m,2,k'}+\sum^{n-1}_{i=2}s_{i,k',i+1,k'},
\]
will be the desired path for this section.

If each point on the trajectory has $d$ possible configurations, the complexity of this method is $O(nd)$.

\subsection{Minimizing the sum of the magnitudes of the joint variations from the starting point to the endpoint}
\label{sec:dijkstra}

The second method is trajectory planning, which minimizes the sum of the magnitudes of the joint variations between each point from the starting point to the endpoint.

For this purpose, we define a weighted graph $G=(V,E)$, 
where $V=\{p_{i,j} \mid i\in\{0,\dots,n\},j\in\{1,\dots,d\}\}$ 
is the set of vertices corresponding to the joint configuration at each point and
$E=\{(p_{i,j},p_{i+1,k})\mid i\in\{0,\dots,n-1\}, j,k\in\{0,\dots,d\}\}$
is the set of edges connecting the vertices at adjacent points.
We estimate the shortest path from the starting point $p_{1,m}$ ($m=1,2,3,4$) to the endpoint $p_{n,k}$ in $G$ using Dijkstra's algorithm \cite{dij1959} as follows, with 
the weight of the edge $(p_{i,j},p_{i+1,k})$ defined as the the magnitude of the joint variations $s_{i,j,i+1,k}$ in \cref{eq:joint-variation}.
As an initial setting, the shortest distance from the starting point to itself is set to $0$, and the shortest distance to all other points is set to $\infty$.

\begin{enumerate}[label=Step \arabic*., leftmargin=13mm]
  \item Set the beginning point to $p_{1,m}$. Set the shortest distance from the starting point to itself to $0$ and the shortest distance to all other points to $\infty$.
  \item Move to the points $p_{2,1},p_{2,2},p_{2,3},p_{2,4}$ which are adjacent to $p_{1,m}$, and compare the existing shortest distances (all $\infty$ for the four points) with the total travel distances to each point ($0+s_{1,m,2,1}=s_{1,m,2,1}$, $0+s_{1,m,2,2}=s_{1,m,2,2}$,
  $0+s_{1,m,2,3}=s_{1,m,2,3}$, $0+s_{1,m,2,4}=s_{1,m,2,4}$), and set the smallest values among 
  $s_{1,m,2,1},s_{1,m,2,2},s_{1,m,2,3},s_{1,m,2,4}$ as the new shortest distances from $p_{1,m}$ to $p_{2,1},p_{2,2},p_{2,3},p_{2,4}$.
  \item For $k=1,2,3,4$, repeat the following calculations: move to the points $p_{3,1},p_{3,2},p_{3,3},p_{3,4}$ which are adjacent to $p_{2,k}$, compare the existing shortest distances with the total travel distances to each point
  \begin{equation}
    \label{eq:213k}
    \begin{split}
      & s_{1,m,2,k}+s_{2,k,3,1},\; s_{1,m,2,k}+s_{2,k,3,2},\\ 
      & s_{1,m,2,k}+s_{2,k,3,3},\; s_{1,m,2,k}+s_{2,k,3,4},
    \end{split}
  \end{equation}
  and set the smallest values among \cref{eq:213k} as the new shortest distances from $p_{1,m}$ to $p_{3,1},p_{3,2},p_{3,3}$, and $p_{3,4}$.
  \item Similarly, for $i=3,\ldots,n-1$ and $k=1,2,3,4$, move to the points $p_{i+1,1}$, 
  $p_{i+1,2}$, $p_{i+1,3}$, $p_{i+1,4}$ which are adjacent to $p_{i,k}$, compare the existing shortest distances with the total travel distances to each point, and set the smaller values as the new shortest distances from $p_{1,m}$ to $p_{i+1,1},p_{i+1,2},p_{i+1,3}$, and $p_{i+1,4}$.
\end{enumerate}
With this process, we have the shortest distances and their corresponding paths (a total of 16) from the starting points $p_{1,1},p_{1,2},p_{13},p_{1,4}$ to the endpoints $p_{n,1},p_{n,2},p_{n,3},p_{n,4}$. 
Then, the path with the shortest distance among them is determined as the desired path.

If each point on the trajectory has $d$ possible configurations, the number of vertices in the graph for one starting point is $1+(n-1)d$, and the number of edges is $d+(n-2)d^2$. For a graph with $V$ vertices and $E$ edges, the computational complexity of calculating the distances between each vertex is estimated to be $O(E)$. The current implementation of Dijkstra's algorithm uses binary heaps \cite{cor-lei-riv-ste2022}, whose computational complexity is $O(\log{V})$. 
Thus, the complexity of Dijkstra's algorithm is estimated to be $O((V+E)\log{V})$,  
and the complexity of this path-planning method for one starting point is estimated by 
\begin{equation}
  O(E)+O((V+E)\log{V})=O(E+(V+E)\log{V}).
\end{equation}
Since $V=1+(n-1)d$ and $E=d+(n-2)d^2$, the complexity of this method is estimated as 
\begin{align*}
  & O(E+(V+E)\log{V}) \\
  &=O(d+(n-2)d^2+(1+nd+(n-2)d^2)\log{(1+(n-1)d)})\\
  &=O(nd^2\log{(nd)}).
\end{align*}

\subsection{Implementation and computational results}

This section describes the implementation and experimental results of the proposed methods.
The implementation of each procedure was based on an implementation \cite{sim2020} in the Python programming language.

For the path used in the experiment, we used the path generated by spline interpolation in 
Tests from 1 to 5 in \Cref{sec:spline-implementation}.
In the experiment, let $T=n=15$, and the number of solutions of the inverse kinematic problem was 
4 ($d=4$) from the start to the end of the path.
The computing environment is as follows: 
Host environment: Intel Core i7-1165G7, RAM 16 GB, Windows 11 Home, VMware Workstation 16 Player,
Guest environment: RAM 13.2 GB, Ubuntu 22.04.3 LTS, Python: 3.10.12.

\Cref{tab:optimal-path-total-angle,tab:optimal-path-total-time} show the results of the experiment.
The columns ``\Cref{sec:optimal-path-between-points}'' and 
``\Cref{sec:dijkstra}'' show the results of the two methods proposed in each section.
\Cref{tab:optimal-path-total-angle} shows the sum of the movements of the joints for each test. 
The result shows that the total change in the joint configuration obtained by the procedure in \Cref{sec:dijkstra} (with Dijkstra's algorithm) is, on average, smaller than that obtained by the procedure in \Cref{sec:optimal-path-between-points}.
\Cref{tab:optimal-path-total-time} shows the computation time for each test.
The average computation time due to the method in 
\Cref{sec:optimal-path-between-points,sec:dijkstra} is the order of $10^{-4}$[s] and $10^{-3}$[s], respectively, which seems to be a practical computing time for both methods, although the method in \Cref{sec:dijkstra} is slightly slower than the method in \Cref{sec:optimal-path-between-points}.

\begin{table}[t]
  \centering
  \caption{The sum of the movements of the joints.}
  \label{tab:optimal-path-total-angle}
  \begin{tabular}{c|cc} 
    \hline
    Test & \Cref{sec:optimal-path-between-points} [rad] & \Cref{sec:dijkstra} [rad]\\
    \hline
    1 & 8.3142 & 4.4013\\ 
    2 & 11.5985 & 9.8986\\ 
    3 & 15.0005 & 13.6731\\ 
    4 & 8.5828 & 6.3277\\ 
    5 & 7.1897 & 5.7108\\ 
    \hline
    Average & 10.1371 & 8.0023 \\ 
    \hline
  \end{tabular}
\end{table}

\begin{table}[t]
  \centering
  \caption{Computing time of the optimal path calculations.}
  \label{tab:optimal-path-total-time}
  \begin{tabular}{c|cc} 
    \hline
    Test & \Cref{sec:optimal-path-between-points} [$10^{-4}$s] & \Cref{sec:dijkstra} [$10^{-4}$s]\\
    \hline
    1 & 1.96 & 39.4\\ 
    2 & 2.16 & 42.8\\ 
    3 & 2.90 & 33.0\\ 
    4 & 2.90 & 40.1\\ 
    5 & 4.44 & 36.4\\ 
    \hline
    Average & 2.87 & 31.3 \\ 
    \hline
  \end{tabular}
\end{table}

\section{Concluding remarks}
\label{sec:conclusion}

In this paper, we proposed an advanced method for solving the inverse kinematics and optimal path and trajectory planning problems for robot manipulators using spline curves and real quantifier elimination based on Comprehensive Gr\"obner Systems (CGS). The proposed method extends previous work by incorporating cubic spline interpolation for smooth path generation and optimizing joint configurations using shortest path algorithms.

The experimental results demonstrated that the proposed method can efficiently generate smooth trajectories and optimize joint configurations. The computation times for optimal path selection seem practical for real-time applications. The results also showed that the method based on Dijkstra's algorithm for minimizing the sum of joint variations outperformed the method based on minimizing joint variations between each point.

Future work includes the following.
For generating smooth paths, our experiments have shown that the spline interpolation method may not always generate feasible paths.
Therefore, it is necessary to develop methods to ensure that the entire path fits within the feasible region. (We have already started working on this issue by the use of B\'esier curves 
\cite{hat-ter-mik2024}.)
In calculating the optimal path, we assumed that the number of solutions of the inverse kinematics problem is the same at each point on the path.
However, in practice, the number of solutions may vary depending on the point or the affine variety of the ideals defined by the inverse kinematic equations.
Thus, investigation will be necessary for determination of the number and continuity of solutions to the inverse kinematics problem (such as detecting singular points).

Furthermore, the proposed method can be applied to other robot manipulators with different degrees of freedom and constraints, and the computational results can be compared with existing methods.

\section*{Acknowledgments}

This work was partially supported by JKA and its promotion funds from KEIRIN RACE.

% \bibliographystyle{sn-mathphys-num}
% \bibliography{casc-2024-myCobot}

% \end{document}


\end{document}